\documentclass[fleqn,10pt]{wlscirep}
\title{Learning to Recognize Actions from Limited Training Examples Using a Recurrent Spiking Neural Model}

\author[1]{Priyadarshini Panda}
\author[2,*]{Narayan Srinivasa}
\affil[1]{School of Electrical \& Computer Engineering, Purdue University, West Lafayette, IN, USA 47907}
\affil[2]{Intel Labs, Hillsboro, OR, USA 97124}

\affil[*]{Correspondence: narayan.srinivasa@intel.com}

\begin{abstract}
A fundamental challenge in machine learning today is to build a model that can learn from few examples. Here, we describe a reservoir based spiking neural model for learning to recognize actions with a limited number of labeled videos. First, we propose a novel encoding, inspired by how microsaccades influence visual perception, to extract spike information from raw video data while preserving the temporal correlation across different frames. Using this encoding, we show that the reservoir generalizes its rich dynamical activity toward signature action/movements enabling it to learn from few training examples. We evaluate our approach on the UCF-101 dataset. Our experiments demonstrate that our proposed reservoir achieves 81.3\%/87\% Top-1/Top-5 accuracy, respectively, on the 101-class data while requiring just 8 video examples per class for training. 
Our results establish a new benchmark for action recognition from limited video examples for spiking neural models while yielding competetive accuracy with respect to state-of-the-art non-spiking neural models.
\end{abstract}
\begin{document}
\flushbottom
\maketitle
%
%
\thispagestyle{empty}


\section*{Introduction}
The exponential increase in digital data with online media, surveillance cameras among others, creates a growing need to develop intelligent models for complex spatio-temporal processing. Recent efforts in deep learning and computer vision have focused on building systems that learn and think like humans \cite{lecun2015deep, mnih2013playing}. Despite the biological inspiration and remarkable performance of such models, even beating humans in certain cognitive tasks \cite{silver2016mastering}, the gap between humans and artificially engineered intelligent systems is still great. One important divide is the size of the required training datasets. Studies on mammalian concept understanding have shown that humans and animals can rapidly learn complex visual concepts from single training examples \cite{lake2011one}. In contrast, the state-of-the-art intelligent models require vast quantities of labeled data with extensive, iterative training to learn suitably and yield high performance. While the proliferation of digital media has led to the availability of massive raw and unstructured data, it is often impractical and expensive to gather annotated training datasets for all of them. This motivates our work on learning to recognize from few labeled data. Specifically, we propose a reservoir based spiking neural model that reliably learns to recognize actions in video data by extracting long-range structure from a small number of training examples. 

Reservoir or Liquid Computing approaches have shown surprising success in recent years on a variety of temporal data based recognition problems (though effective vision applications are still scarce) \cite{maass2010liquid, lukovsevivcius2009reservoir, srinivasa2014unsupervised}. This can be attributed to populations of recurrently connected spiking neurons, similar to the mammalian neocortex anatomy \cite{wehr2003balanced}, which create a high-dimensional dynamic representation of an input stream. The advantage with such an approach is that the reservoir (with sparse/random internal connectivity) implicitly encodes the temporal information in the input and provides a unique non-linear description for each input sequence, that can be trained for read out by a set of linear output neurons. In fact, in our view, the non-linear integration of input data by the high-dimensional dynamical reservoir results in generic stable internal states that generalize over common aspects of the input thereby allowing rapid learning from limited examples. While the random recurrent connectivity generates favorable complex dynamic activity within a reservoir, it poses unique challenges that are generally not encountered in constructing feedforward/deep learning networks. 

A pressing problem with recurrently connected networks is to determine the connection matrix that would be suitable for making the reservoir perform well on a particular task, because it is not entirely obvious how individual neurons should spike while interacting with other neurons. To address this, Abbott et. al \cite{abbott2016building}, proposed an elegant method for constructing recurrent spiking networks, termed as Autonomous (A) models, from continuous variable (rate) networks, called Driven (D) models. The basic idea in the D/A approach is to construct a network (Auto) that performs a particular task by copying another network (Driven) statistics that does the same task. The copying provides an estimate of the internal dynamics for the autonomous network including currents at individual neurons in the reservoir to produce the same outputs. Here, we adopt the D/A based reservoir construction approach and modify it further by introducing approximate delay apportioning to develop a recurrent spiking model for practical action recognition from limited video data. Please note, we use Auto/Autonomous interchangeably in the remainder of the text.

Further, in order to facilitate limited example training in a reservoir spiking framework, we propose a novel spike based encoding/pre-processing method to convert the raw pixel valued videos in the dataset into spiking information that preserves the temporal statistics and correlation across different frames. This approach is inspired by how MicroSaccades (MS) cause retinal ganglion cells to fire synchronously corresponding to contrast edges after the MS. The emitted synchronous spike volley thus rapidly transmits the most salient edges of the stimulus, which often constitute the most crucial information \cite{masquelier2016microsaccades}. In addition, our encoding eliminates irrelevant spiking information due to ambiguity such as, noisy background activity or jitter (due to unsteady camera movement), further making our approach more robust and reliable. Hence, our encoding, in general, captures the signature action or movement of a subject across different videos as spiking information. This, in turn, enables the reservoir to recognize/generalize over motion cues from spiking data, to enable learning various types of actions from few video samples per class. Our proposed spiking model is much more adept at learning object/activity types, due in part to our ability to analyze activity dynamically as it takes place over time.


\section*{Reservoir Model: Framework \& Implementation}
\subsection*{Model architecture}
Let us first discuss briefly the Driven/Autonomous model approach \cite{abbott2016building}. The general architecture of the Driven/      
Autonomous reservoir model is shown in Fig. 1 (a). Both  driven/auto models are 3-layered networks. The input layer is connected to a reservoir that consists of nonlinear (Leaky-Integrate-and-Fire, LIF), spiking neurons recurrently connected to each other in a random sparse manner (generally with a connection probability of 10\%), and the spiking output produced by the reservoir is then projected to the output/readout neurons. In case of driven, the output neurons are rate neurons that simply integrate the spiking activity of the reservoir neurons. On the other hand, the auto model is an entirely spiking model with LIF neurons at the output as well as within the reservoir. Please note, each reservoir neuron here is excitatory in nature and the reservoir does not have any inhibitory neuronal dynamics in a manner consistent with conventional Liquid or Reservoir Computing.

Essentially, the role of the driven network is to provide targets ($f_{out}$) for the auto model, which is the spiking network we are trying to construct for a given task. To achieve this, the driven network is trained on an input ($f_D$) that is a high pass filtered version of the desired output ($f_{out}$), 

\begin{equation}
f_D = f_{out} + \tau_{fast} \frac{df_{out}}{dt}
\end{equation}

The connections from the reservoir to the readout neurons ($w$) of the driven model are then trained to produce desired activity using the Recursive Least Square (RLS) rule \cite{haykin2008adaptive, sussillo2009generating}. Once trained, the driven network connections ($J_{fast}$) are imported into the autonomous model and a new set of network connections ($J_{slow}$) are added for each fast connection. The combination of the two sets of synapses, $J_{fast}$ and $J_{slow}$, allows the auto model to match its internal dynamics to that of the driven network that is already tuned to producing desired $f_{out}$. In fact, the connections $u$ and $J_{slow}$ in the auto model are derived from the driven network as, $u = u_Du_R$ and $J = u_DBw$, where $u_R$ and $B$ are randomly chosen. Note, the time constants of the two sets of synapses is different i.e. $\tau_{slow} \sim 10*\tau_{fast}$ that allows the autonomous network to slowly adjust its activity to produce $f_{out}$ during training. For further clarification, please refer to \cite{abbott2016building} to gain more details and insights on D/A approach. After the autonomous model is constructed, we train its output layer (that consist of LIF spiking neurons), $w’$, using a supervised Spike Timing Dependent Plasticity (STDP) rule \cite{ponulak2010supervised} based on the input $f_{in}$. 

\begin{figure}[h]
\centering
\includegraphics[width=\linewidth]{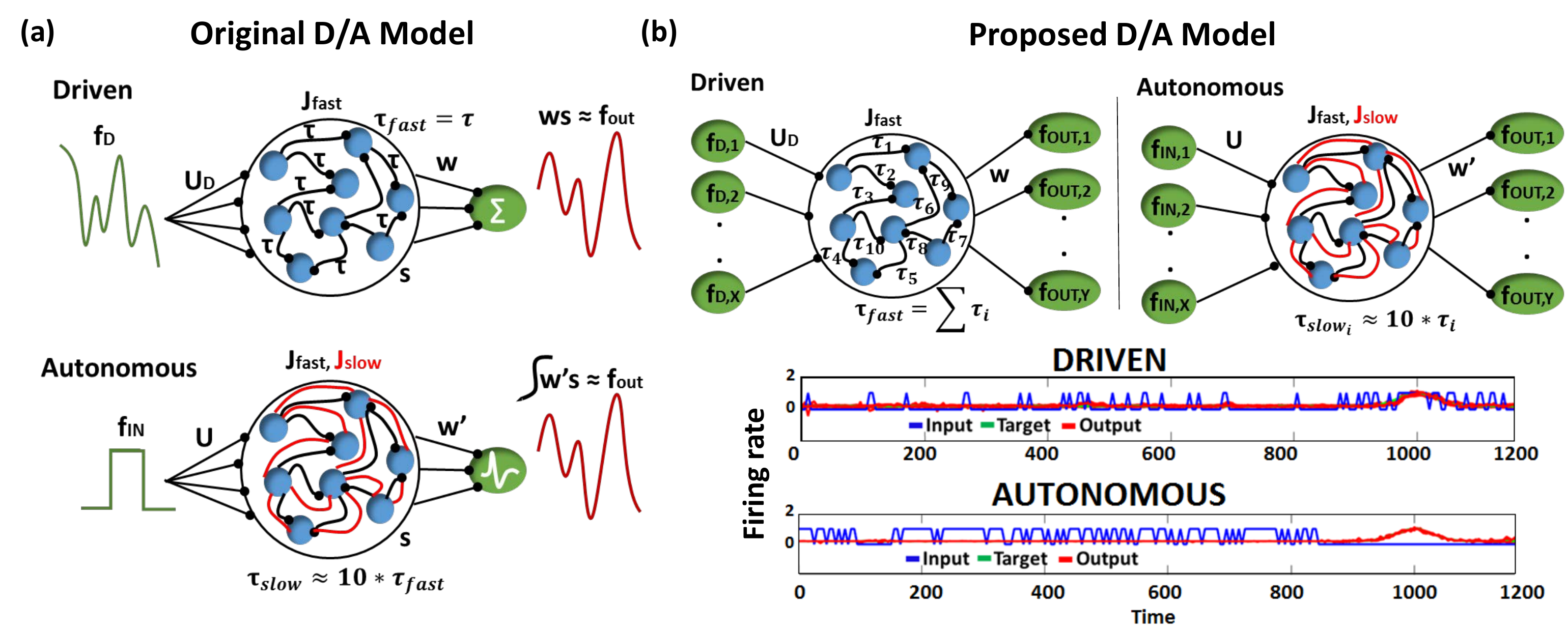}
\caption{(a) Structure of Driven (Autonomous) networks with rate (spiking) output neurons and fast (fast-slow) synaptic connections. Here, fast connections have equivalent delay time constant. (b) (Top) Structure of Driven-Autonomous model with the proposed  $\tau_{fast}$ apportioning to construct variable input(denoted as $X$)/output (denoted as $Y$) reservoir models. Here, each fast connection has a different delay time constant. (Bottom) Desired output produced by an autonomous model and its corresponding Driven model, that is constructed using approximate $\tau_{fast}$ apportioning. Please note, input spiking activity (blue curve) and output (red/green curve) in driven are similar since input is a filtered version of the output, while autonomous model works on random input data.}
\label{fig1}
\end{figure}

The motif behind choosing $f_D$ as a filtered version of $f_{out}$ is to compensate for the synaptic filtering at the output characterized by time constant, $\tau_{fast}$ \cite{eliasmith2005unified, abbott2016building}. However, it is evident that the dependence of $f_D$ on $f_{out}$ from Eqn. 1 imposes a restriction on the reservoir construction, that is, the number of input and output neurons must be same. To address this limitation and extend the D/A construction approach to variable input-output reservoir models, we propose an approximate definition of the driven input using a delay apportioning method as,

\begin{equation}
{({f_D}_i)}_j = {(f_{out})}_j + {\tau_{fast}}_i \frac{{(df_{out})}_j}{dt} 
\end{equation}

Here, $i$ varies from \textit{1 to X}, $j$ varies from \textit{1 to Y}, where $X/Y$ are the number of input/output neurons in our reservoir model respectively and $\tau_{fast} = \sum_{i=1}^{X} {\tau_{fast}}_i$. Fig. 1 (b, Top) shows the architecture of the D/A model with the proposed delay apportioning. For each output function (${(f_{out})}_j$), we derive the driven inputs ${({f_D}_{i=1...X})}_j$ by distributing the time constant across different inputs. We are essentially compensating the phase delay due to $\tau_{fast}$ in an approximate manner. Nevertheless, it turns out this approximate version of $f_D$ still drives the driven network to produce the desired output and  derive the connection matrix of the autonomous model that is then trained on the real input (video data in our case). Fig. 1 (b, Bottom) illustrates the desired output genereated by a D/A model of 2(input)$\times$400(reservoir neurons)$\times$1(output) configuration constructed using Eqn. 2 . Note, for the sake of convenience in representation and comparative purposes, the output of both driven/auto models are shown as rate instead of spiking activity. It is clearly seen in the driven case that the input spiking activity for a particular input neuron ${f_D}_i$, even with $\tau_{fast}$ apportioning, approximates the target quite well. Also, the autonomous model derived from the driven network in Fig. 1 (b, Bottom), produces the desired spiking activity in response to random input upon adequate training of $w'$ connections. 

 \subsection*{Supervised STDP \& One-Hot Encoding}
 \begin{figure}[h]
\centering
\includegraphics[width=0.75\linewidth]{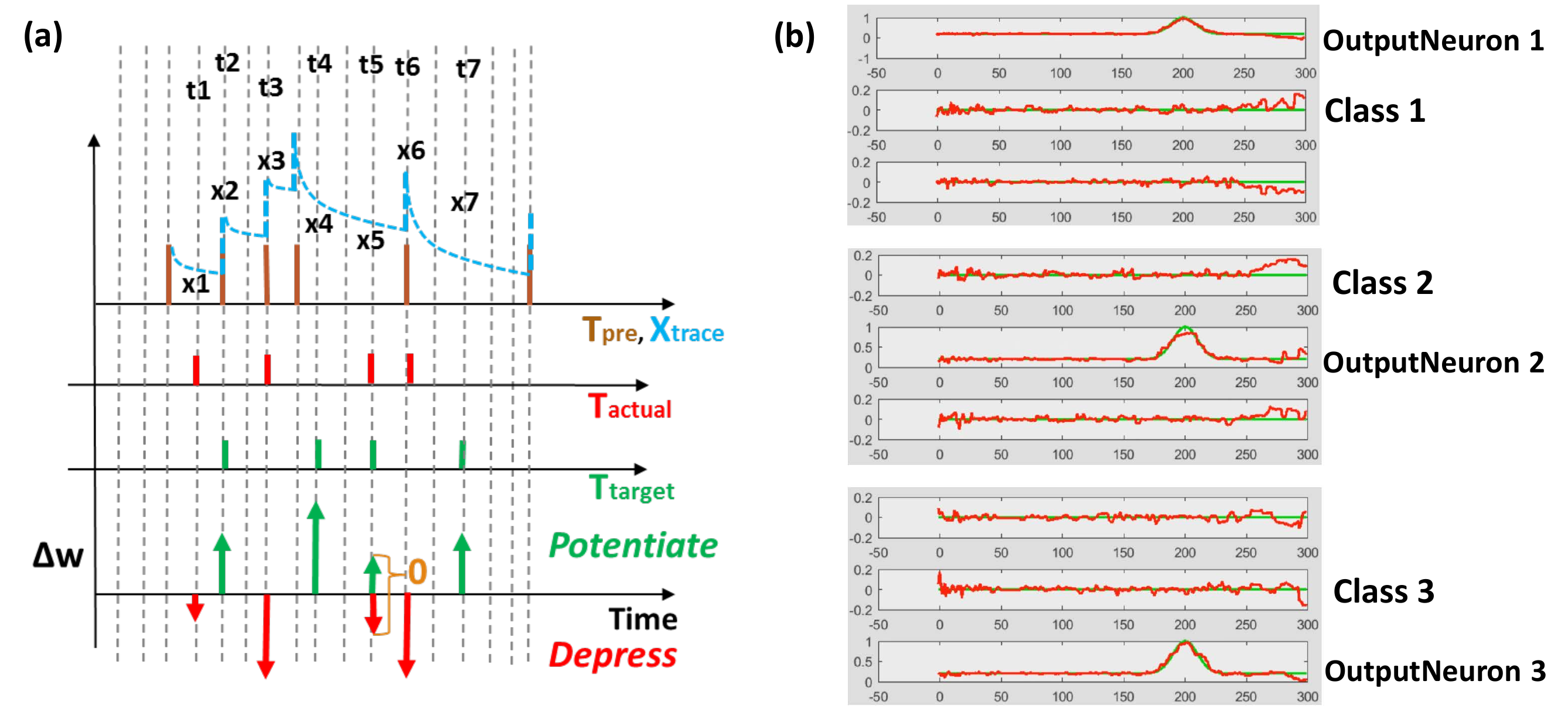}
\caption{(a) Supervised STDP rule (defined by Eqn. 3) using pre-synaptic trace to perform potentiation/depression of weights. $t_1, t_2...t_7$ are the time instants at which target/actual spiking activity occurs resulting in a weight change. The extent of potentiation or depression of weights $\Delta w$ is proportional to the value of the pre-synaptic trace $x_1, x_2...x_7$ at a given time instant. (b) One-hot spike encoding activity for a 3-class problem: Based on the class of the training input, the corresponding output neuron is trained to produce high activity while remaining neurons yield near-zero activity.}
\label{fig2}
\end{figure}

Reservoir framework relaxes the burden of training by fixing the connectivity within the reservoir and that from input to the reservoir. Merely, the output layer of readout neurons ($w, w'$ in Fig. 1 (a)) is trained (generally in a supervised manner) to match the reservoir activity with the target pattern. In our D/A based reservoir construction, readout of the driven network is trained using standard RLS learning \cite{sussillo2009generating} while we use a modified version of the supervised STDP rule proposed in \cite{ponulak2010supervised} to conduct autonomous model training. An illustration of our STDP model is shown in Fig. 2 (a). For a given target pattern, the synaptic weights are potentiated or depressed as

\begin{equation}
\Delta w = x_{trace} (t_{target} - t_{actual})
\end{equation}

Where $t_{actual}/t_{target}$ denotes the time of occurrence of actual/desired spiking activity at the post-synaptic neuron during the simulation and $x_{trace}$, namely the presynaptic trace, models the pre-neuronal spiking history. Every time a pre-synaptic neuron fires, $x_{trace}$,  increases by 1, otherwise  it decays exponentially with $\tau_{pre}$ that is of the same order as $\tau_{fast}$. Fundamentally, as illustrated in Fig. 2 (a), Eqn. 3 depresses (or potentiates) the weights at those time instants when actual (or target) spike activity occurs. This ensures that actual spiking converges toward the desired activity as training progresses. Once the desired/actual activity become similar, the learning stops, since at time instants where both desired and actual spike occur simulataneously, the weight update value as per Eqn. 3 becomes 0. 

Since the main aim of our work is to develop a spiking model for action recognition in videos, we model the desired spiking activity at the output layer based on one-hot encoding. As with standard machine learning applications, the output neuron assigned to a particular class is trained to produce high spiking activity while the remaining neurons are trained to generate zero activity. Fig. 2 (b) shows the one-hot spike encoding for a 3-class problem, wherein the reservoir’s readout has 3 output neurons and based on the training input’s class, the desired spiking activity (sort of ‘bump’) for each output neuron varies. Again, we use the supervised STDP rule described above to train the output layer of the autonomous model. Now, it is obvious that since the desired spiking activity of the auto model ($f_{out}$ referring to Fig. 1) follows such one-hot spike encoding, the driven network (from which we derive our auto model) should be constructed to produce equivalent target output activity. To reiterate, the auto model (entirely spiking) works on the real-world input data and is trained to perform required tasks such as classification. The connection matrix ($J_{slow}$) of the auto model is derived from a driven network (with continuous rate-based output neurons) that, in turn, provides the target activity for the output neurons of the auto model. 

\section*{Input Processing}
In this work, we use the UCF101 dataset \cite{soomro2012ucf101} to perform action recognition with our reservoir approach.  UCF101 consists of 13320 realistic videos (collected from YouTube) that fall into 101 different categories or classes, 2 of which are shown in Fig. 3. Since the reservoir computes using spiking inputs, we convert the pixel valued action information to spike data in steps outlined below.

\subsection*{Pixel Motion Detection}
\begin{figure}[h]
\centering
\includegraphics[width=0.75\linewidth]{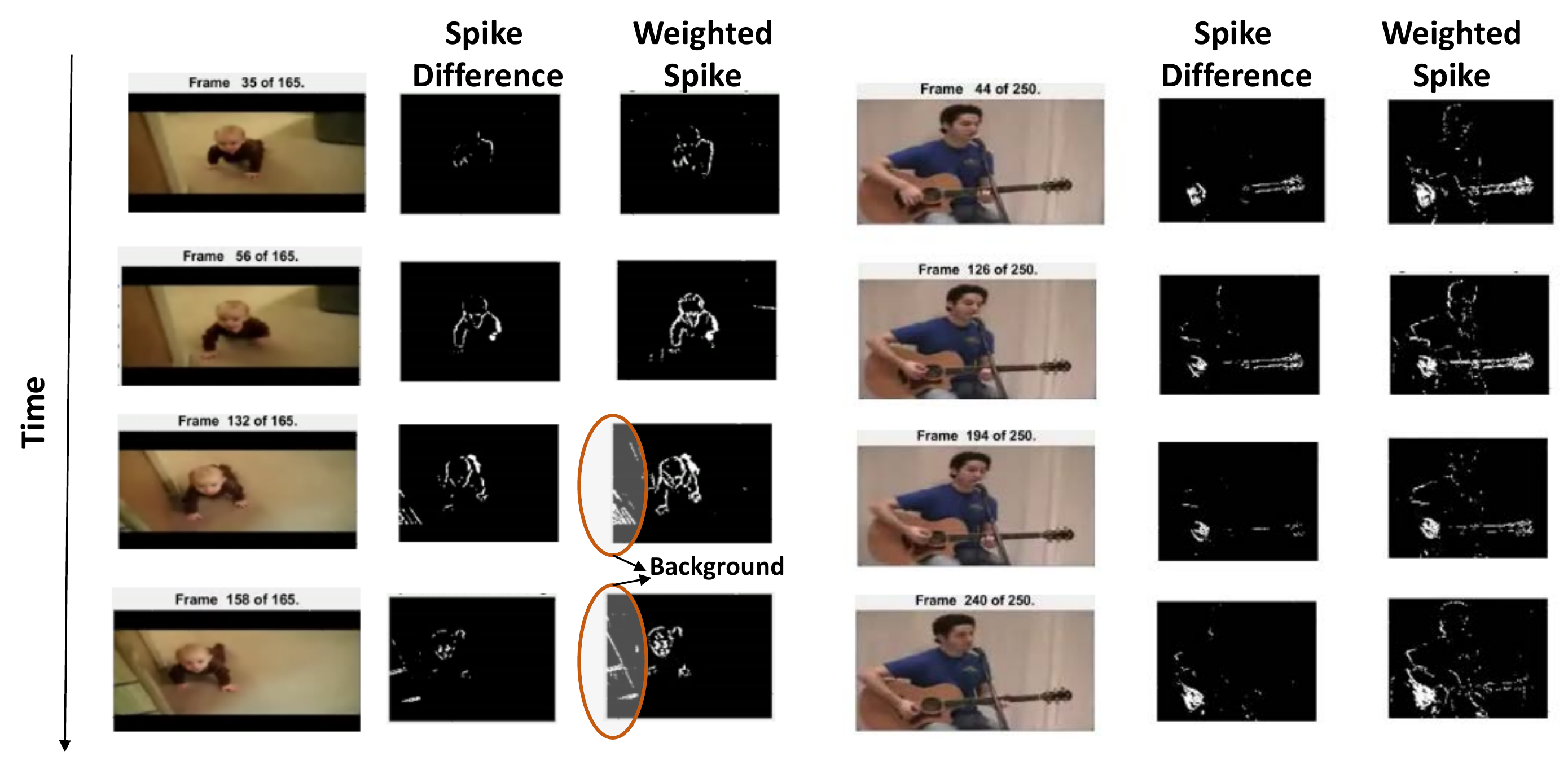}
\caption{Fixed threshold spiking input (denoted by Spike Difference) and variable threshold weighted input data (denoted by Weighted spike) shown for selected frames for two videos of the UCF101 dataset. As time progresses, the correlation across spiking information is preserved. Encircled regions show the background captured in the spiking data. }
\label{fig3}
\end{figure}

Unlike static images, video data are a natural target in event-based learning algorithms due to their implicit time-based processing capability. First, we track the moving pixels across different frames so that the basic action/signature is captured. This is done by monitoring the difference of the pixel intensity values ($P_{diff}$) between consecutive frames and comparing the difference against some threshold ($\epsilon$, that is user-defined) to detect the moving edges (for each location (x,y) within a frame), as  

\begin{equation}
\begin{split}
P_{diff}(x,y) = Frame_j-Frame_{j-1} ; (\textit{where j = 2 to NumberOfFrames}) \\
\textbf{If} \hspace{2mm} P_{diff}(x,y) \ge \epsilon, \hspace{2mm} Spike_{diff}(x,y) =1\\
WSpike_{diff}(x,y) = \frac{\sum_i \epsilon_i Spike_{diff}(x,y)}{\sum_i \epsilon_i}; (\textit{where i=1 to N}) \\
\end{split}
\end{equation}

The thresholding detects the moving pixels and yields an equivalent spike pattern ($Spike_{diff}$) for each frame. It is evident that the conversion of the moving edges into spikes preserves the temporal correlation across different frames. While $Spike_{diff}$ encodes the signature action adequately, we further make the representation robust by varying the threshold \textit{($\epsilon_i$ = \{1,2,4,8,16,32\}; N=6 in our experiments)} and computing the weighted summation of the spike differences for each threshold as shown in the last part of Eqn. 4. The resultant weighted spike difference $WSpike_{diff}(x,y)$ is then binarized to yield the final spike pattern for each frame. This kind of pixel-change based motion detection to generate spike output is now possible with a new breed of energy efficient sensors inspired by the retina \cite{DVSfnins}.

Fig. 3 shows the spiking inputs captured from the moving pixels based on fixed threshold spike input and varying threshold weighted input for selected frames. It is clearly seen that the weighted input captures more relevant edges and even subtle movements, for instance, in the PlayingGuitar video, the delicate body movement of the person is also captured along with the significant hand gesture. Due to the realistic nature of the videos in UCF101, there exists a considerable amount of intra class variation along with ambiguous noise. Hence, a detailed input pattern that captures subtle gestures facilitates the reservoir in differentiating the signature motion from noise. It should be noted that we convert the RGB pixel video data into grayscale before performing motion detection.

\subsection*{Scan based Filtering}
We observe from Fig. 3, for the BabyCrawling case, that along with the baby’s movement, some background activity is also captured. This background spiking occurs on account of the camera movement or jitter. Additionally, we see that majority of the pixels across different frames in Fig. 3 are black since the background is largely static and hence does not yield any spiking activity. To avoid the wasteful computation due to non-spiking pixels as well as circumvent insignificant activity due to camera motion, we create a Bounding Box (BB) around the Centre of Gravity (CoG) of spiking activity for each frame and scan across 5 directions as shown in Fig. 4. The CoG ($x_g, y_g$) is calculated as the average of the active pixel locations within the BB. In general, the dimensionality of the video frames in the UCF101 dataset is $\sim 200\times300$. We create a BB of size $41\times41$ centered at the CoG, capture the spiking activity enclosed within the BB, in what we refer to as the Center (C) scan, and then stride the BB along different directions (by number of pixels equal to $\sim$ half the distance between the CoG and the edge of the BB), namely, Left (L), Right (R), Top (T), Bottom (B), to capture the activity in the corresponding directions. These five windows of spiking activity can be interpreted as capturing the salient edge pixels (derived from pixel motion detection above) after each microsaccade \cite{masquelier2016microsaccades}. In our model, the first fixation is at the center of the image. Then, the fixation is shifted from the center of the image to the CoG using the first microsaccade. The rest of the four microsaccades are assumed to occur in four directions (i.e., top, bottom, left and right) with respect to CoG of the image by the stride amount. It should be noted that this sequence of four miscrosaccades is assumed to be always fixed in its direction. The stride amount can, however, vary based on the size of the BB.

Fig. 4 (a) clearly shows that the CRLTB scans only retain the relevant region of interest across different frames of the BabyCrawling video while filtering out the irrelevant spiking activity. Additionally, we also observe that in the PlayingGuitar video shown in Fig. 4 (b), different scans capture diverse aspects of the relevant spiking activity, for instance, hand gesture in C scan while guitar shape in B scan shown in the 3rd row. This ensures that all significant movements for a particular action in a given frame are acquired in a holistic manner.

\begin{figure}[h]
\centering
\includegraphics[width=\linewidth]{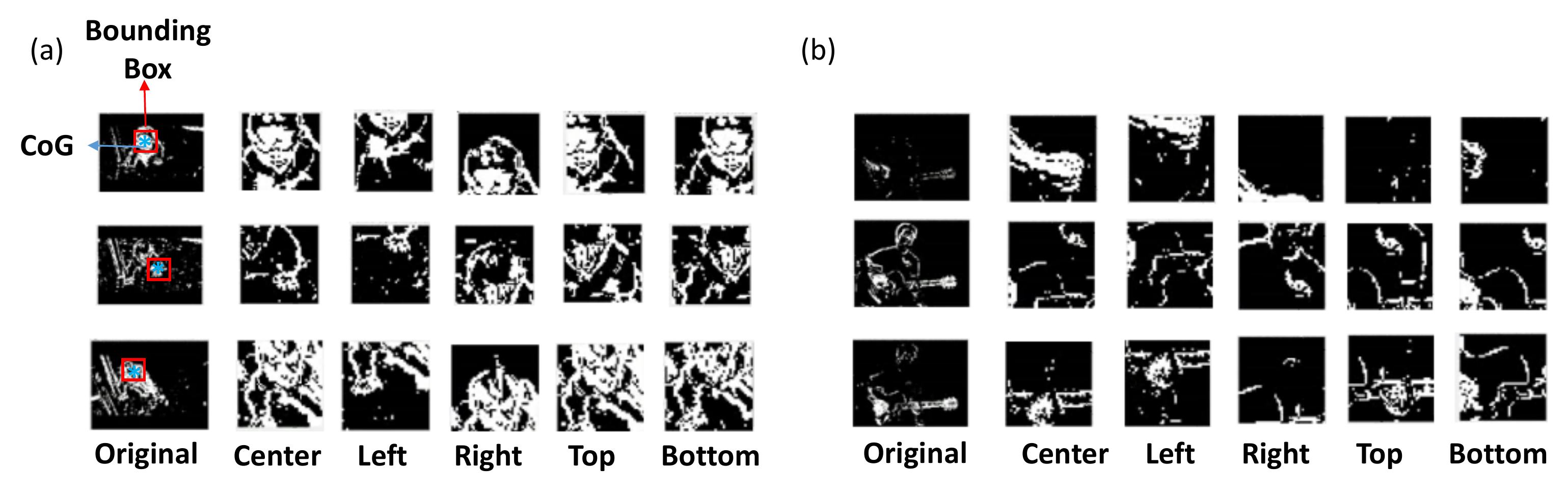}
\caption{Scan based filtering from the original spike data resulting in Center, Left, Right, Top, Bottom scans per frame for (a) BabyCrawling video (b) PlayingGuitar video. Each row shows the original frame as well as the different scans captured.}
\label{fig4}
\end{figure}

In addition to the scan based filtering, we also delete certain frames to further eliminate the effect of jitter or camera movement on the spike input data. We delete the frames based on two conditions:
\begin{itemize}
\item{If the overall number of active pixels for a given spiking frame are greater/lesser than a certain maximum/minimum threshold, then, that frame is deleted. This eliminates all the frames that have flashing activity due to unsteady camera movement. In our experiments, we use a min/max threshold value of 8\%/75\% of the total number of pixels in a given frame. Note, we perform this step before we obtain the CRLTB scans.}
\item{Once the CRLTB scans are obtained, we monitor the activity change for a given frame in RLTB as compared to C (i.e. for $Frame_i$ check $C_i - \{R_i , L_i , T_i , B_i\}$) . If the activity change (say, $C_i -R_i$) is greater than the mean activity across $C_i$, then, we delete that particular frame ($R_i$) corresponding to the scan that yields high variance. 
}
\end{itemize}

It is noteworthy to mention that the above methods of filtering eliminate the ambiguous motion or noise in videos with minimal background activity or jitter motion (such as PlayingGuitar) to a large extent. 
However, the videos where the background has significant motion (such as HorseRacing) or those where multiple human subjects are interacting (such as Fencing, IceDancing), the extracted spike pattern does not encode a robust and reliable signature motion due to extreme variation in movement. 
For future implementations, we are studying using a DVS camera \cite{li2017cifar10, DVSfnins} to extract reliable spike data despite ego motion. Nonetheless, our proposed input processing method yields reasonable accuracy (shown later in Results section) with the D/A model architecture for action recognition with limited video examples. 
Supplementary Video 1,2 give better visualization of the spiking data obtained with the pixel motion detection and scan based filtering scheme.

\section*{Experimental Methodology}
\subsection*{Network Parameters}
The proposed reservoir model and learning was implemented in MATLAB. The dynamics of the reservoir neurons in the Driven/Autonomous models are described by
\begin{equation}
\begin{split}
\tau_{mem} \frac{dV}{dt} = V_{rest} - V + (J_{fast} f(t) + u_D f_D) ; \hspace{2mm} Driven\\
\tau_{mem} \frac{dV}{dt} = V_{rest} - V + (J_{fast} f(t) +J_{slow} s(t) + u f_{in}) ; \hspace{2mm} Auto\\
\end{split}
\end{equation}
where $\tau_{mem} =$ \textit{20 ms}, $f_{in}$ denotes the real input spike data, $f_D$ is the derived version of the target output (refer to Eqn. 2), $u_D /u$ are the weights connecting the input to the reservoir neurons in the driven/auto model respectively. Each neuron fires when its membrane potential, $V$, reaches a threshold $V_{th} =$ \textit{-50 mV} and is then reset to $V_{reset} = V_{rest} =$ \textit{-70 mV}. Following a spike event, the membrane potential is held at the reset value $V_{reset}$ for a refractory period of \textit{5 ms}. The parameters $f(t), s(t)$ correspond to fast/slow synaptic currents observed in the reservoir due to the inherent recurrent activity. In the driven model, only fast synaptic connections are present while auto model has both slow/fast connections. The synaptic currents are encoded as traces, that is, when a neuron in the reservoir spikes, $f(t)$ (and $s(t)$ in case of auto model) is increased by 1, otherwise it decays exponentially as
\begin{equation}
\begin{split}
\tau_{slow} \frac{ds(t)}{dt} = -s(t); \hspace{6mm}
\tau_{fast} \frac{df(t)}{dt} = -f(t)\\
\end{split}
\end{equation}

where $\tau_{slow} =$ \textit{100 ms}, $\tau_{fast} =$ \textit{5 ms}. Both $J_{slow}$ and $J_{fast}$ recurrent connections within the reservoir of D/A model are fixed during the course of training. $J_{fast}$ of the driven model are randomly drawn from a normal distribution. $J_{slow}$ and $u$ (refer to Fig. 1 (a)) of the auto network are derived from the driven model using \textit{$u_R$ = 1, B = random number drawn from a uniform distribution in the range [0, 0.15]}. We discuss the impact of variation of $B$ on the decoding capability of the autonomous reservoir in a later section. If the reservoir outputs a spike pattern, say $x(t)$, the weights from the reservoir to the readout ($w$) are trained to produce the desired activity, $f_{out}$. The continuous rate output neurons in driven network are trained using RLS such that $w*x(t) \sim f_{out}(t)$. The spiking output neurons of the auto model (with similar neuronal parameters as Eqn. 5) integrate the net current $w'*x(t)$ to fire action potentials, that converge toward the desired spiking activity with supervised STDP learning. 

\subsection*{D/A model for Video Classification}
The input processing technique yields 5 different CRLTB scan spike patterns per frame for each video. In the D/A based reservoir construction, each of the 5 scans for every video is processed by 5 different autonomous models. Fig. 5 shows an overview of our proposed D/A construction approach for video classification. Each scan is processed separately by the corresponding autonomous model. An interesting observation here is that since the internal topology across all auto models is equivalent, they can all be derived from a single driven model as shown in Fig. 5. In fact, there is no limit to the number of autonomous models that can be extracted from a driven network, which is the principal advantage of our approach. Thus, in case, more scans based on depth filtering or ego motion monitoring are obtained during input processing, our model can be easily extended to incorporate more autonomous models corresponding to the new scans.

\begin{figure}[h]
\centering
\includegraphics[width=0.75\linewidth]{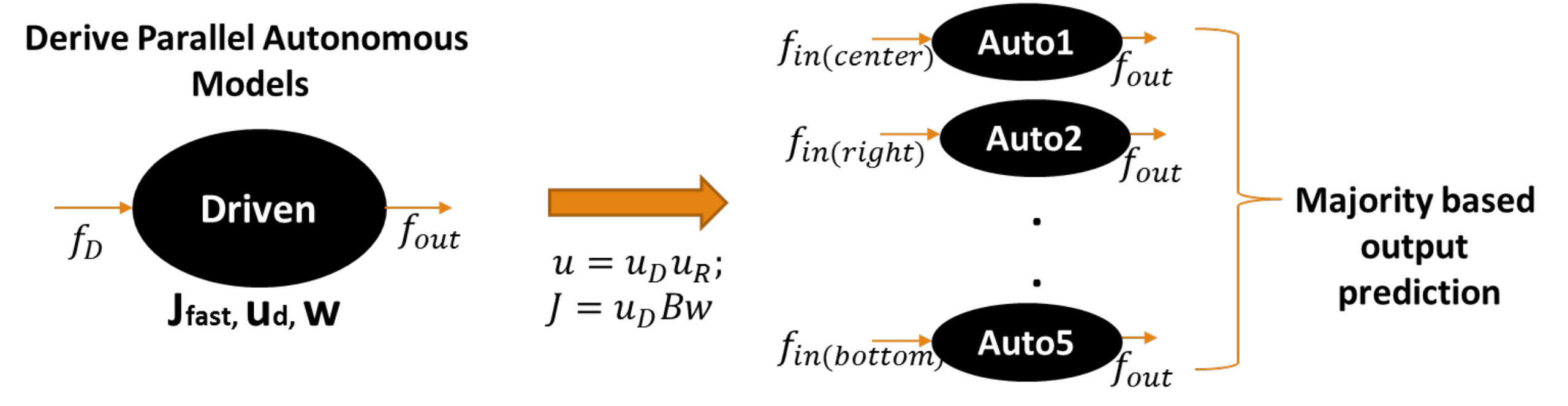}
\caption{Proposed Driven/Autonomous construction scheme for processing the CRLTB scans obtained from the scan based filtering}
\label{fig5}
\end{figure}

During training, the driven network is first trained to produce the target pattern (similar to the one-hot spike encoding described in Fig. 2 (b)). Then, each autonomous model is trained separately on the corresponding input scan to produce the target pattern. In all our experiments, we train each auto model for 30 epochs. In each training epoch, we present the training input patterns corresponding to all classes sequentially (for instance, Class1$\rightarrow$Class 2$\rightarrow$Class 3) to produce target patterns similar to Fig. 2 (b). Each input pattern is presented for a time period of \textit{300 ms} (or 300 time steps) . In each time step, a particular spike frame for a given scan is processed by the reservoir. For patterns where total number of frames is less than 300, the network still continues to train (even in absence of input activity) on the inherent reservoir activity generated due to the recurrent dynamics. In fact, this enables the reservoir to generalize its' dynamical activity over similar input patterns belonging to the same class while discriminating patterns from other classes. Before presenting a new input pattern, the membrane potential of all neurons in the reservoir are reset to their resting values.

After training is done, we pass the test instance CRLTB scans to each auto model simultaneously and observe the output activity. The test instance is assigned a particular class label based on the output neuron class that generated the highest spiking response. This assignment is done for each auto model. Finally, we take a majority vote across all the assignments (or all auto models) for each test input to predict its class. For instance, if majority of the auto models (i.e. $\ge 3$) predict a class (say Class 1), then predicted output class from the model is Class 1. Now, a single auto model might not have sufficient information to recognize the test input correctly. The remaining scans or auto models, in that case, can compensate for the insufficient or inaccurate output prediction from the individual models. While we reduce the computational complexity of training by using the scan based filtering approach that partitions the input space into smaller scans, the parallel voting during inference enhances the collective decision making of the proposed architecture, thereby improving the overall performance. 

The reservoir (both driven/auto) topology in case of the video classification experiments consist of $X(InputNeurons)-N(ReservoirNeurons)-Y(OutputNeurons)$ with $p \%$ connectivity within the reservoir. $X$ is equivalent to the size of the bounding box used during scan based filtering (in our case ${X = 41\times41 =1681}$). The number of output neurons vary based on the classification problem, that is, $Y = 5$ for a 5-class problem. In most experiments below, we use $p =10\%$ connectivity to maintain the sparseness in the reservoir. In the Results section below, we further elaborate on the role of connectivity on the performance of the reservoir. The number of neurons $N$ are varied based upon the complexity of the problem in order to achieve maximum classification accuracy. 
 
\begin{figure}[h]
\centering
\includegraphics[scale =0.5]{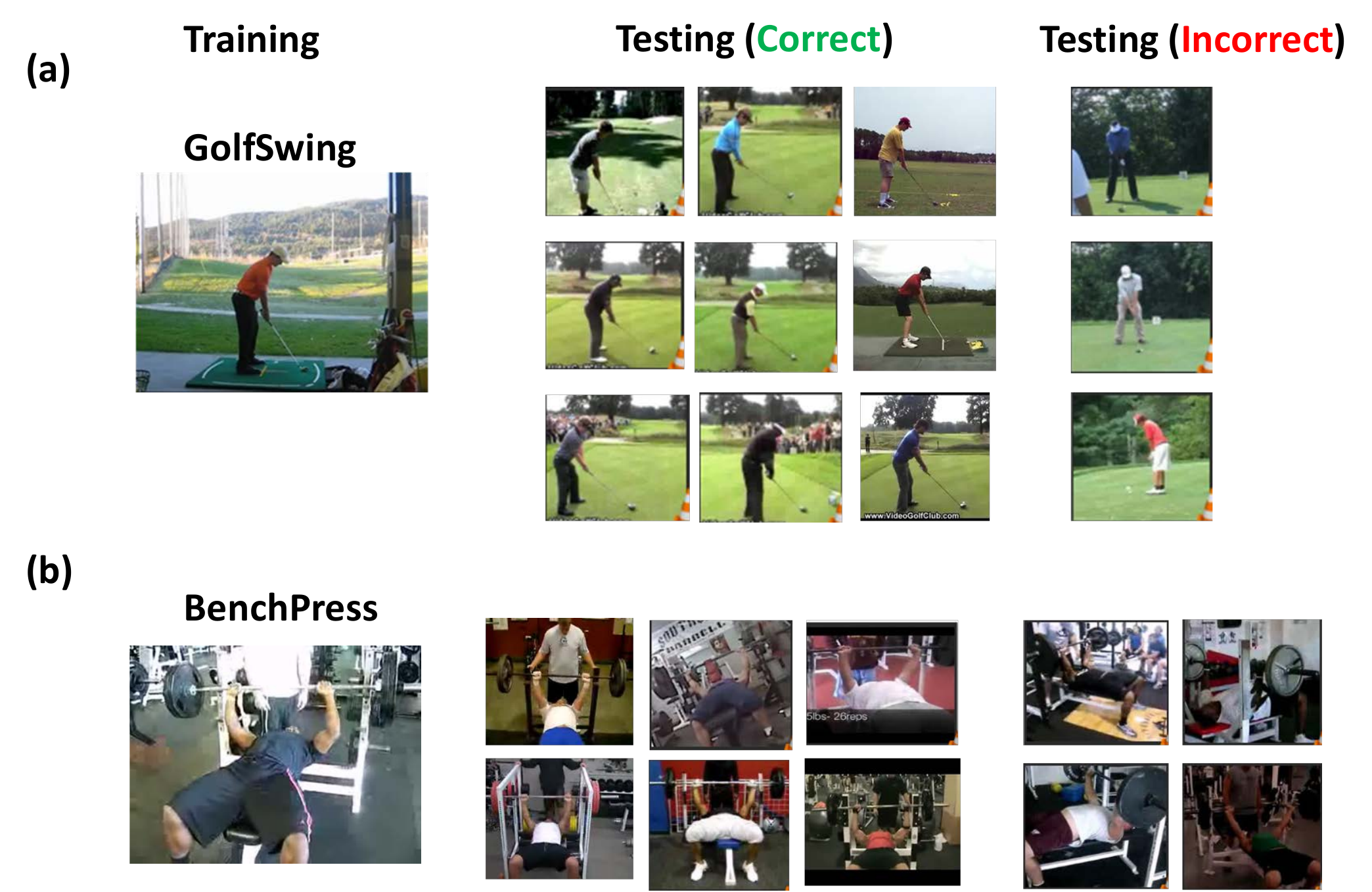}
\caption{Training/Testing video clips of the 2-class problem (a) GolfSwing (b) BenchPress: Column 1 shows the training video, Column 2/3 show the test video examples correctly/incorrectly classified with our D/A model.}
\label{fig6}
\end{figure}
 
\subsection*{Classification Accuracy on UCF101}
First, we considered a 2-class problem of recognizing BenchPress/GolfSwing class in UCF101 to test the effectiveness of our overall approach. We specifically chose these 2 classes in our preliminary experiment since majority of videos in these categories had a static background and the variation in action across different videos was minimal. We simulated an 800N-2Output, 10\% connectivity D/A reservoir (N:number of reservoir neurons) and trained the autonomous models with a single training video from each class as shown in Fig. 6. It is clearly seen that even with single video training, the reservoir correctly classifies a host of other examples irrespective of the diverse subjects performing the same action. For instance, in the golf examples in Fig. 6 (a), the reservoir does not discriminate between a person wearing shorts or trousers while recognizing the action. Hence, we can gather that the complex transient dynamics within a reservoir latches onto a particular action signature (swing gesture in this case) that enables it to classify several test examples even under diverse conditions. Investigating the incorrect predictions from the reservoir for both classes, we found that the reservoir is sensitive toward variation in views and inconsistencies in action, which results in inaccurate prediction. Fig. 6 shows that the incorrect results are observed when the view angle of the training/test video changes from frontal to adjacent views. In fact, in case of BenchPress, even videos with similar view as training are incorrectly classified due to disparity in motion signature (for instance, number of times the lifting action is performed). Note, we only show a selected number of videos classified correctly/incorrectly in Fig. 6. 

To account for such variations, we increased the number of training examples from \textit{1 to 8}. In the extended training set, we tried to incorporate all kinds of view/signature based variations for each action class as shown in Fig. 7 (a). For the 2-class problem, with 8 training examples for each class, we simulated a 2000N-10\% connectivity reservoir that yields an accuracy of 93.25\%. Fig. 7 (b) shows the confusion matrix for the 2-class problem. BenchPress has more misses than GolfSwing as the former has more variation in action signature across different videos. Now, it is well-known that the inherent chaotic or spontaneous activity in a reservoir enables it to produce a wide variety of complex output patterns in response to an input stimulus. In our opinion, this behavior enables data efficient learning as the reservoir learns to generalize over a small set of training patterns and produce diverse yet converging internal dynamics that transform different test inputs toward the same output. Note, testing is done on all the remaining videos in the dataset that were not selected for training. 

\begin{figure}[h]
\centering
\includegraphics[width=\linewidth]{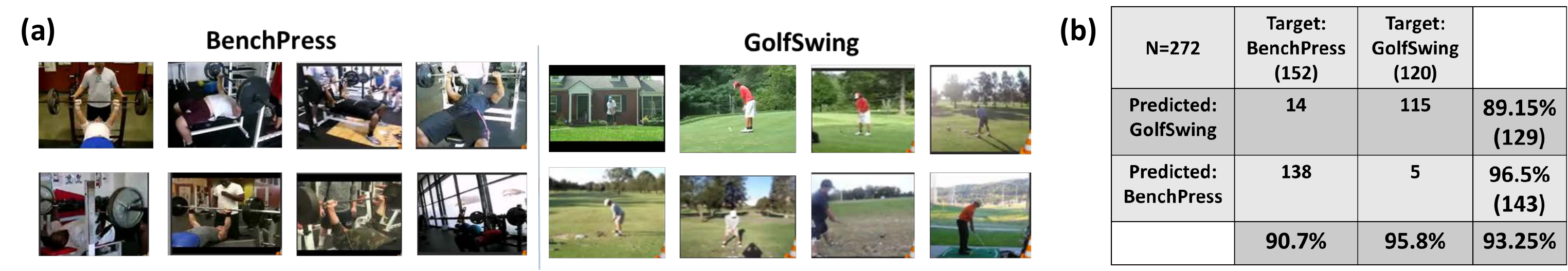}
\caption{(a) Clips of the 8 training videos for each class (BenchPress/GolfSwing) incorporating multiple views/variation in action signature (b) Confusion Matrix showing the overall as well as class-wise accuracy for 2-class BenchPress/GolfSwing classification}
\label{fig7}
\end{figure}

Next, we simulated a reservoir of 4K/6K/8K/9K neurons (with 10\% connectivity) to learn the entire 101 classes in the UCF101 dataset. Note, for each of the 101 classes, we used 8 different training videos that were randomly chosen from the dataset. To ensure that more variation is captured for training, we imposed a simple constraint during selection, that each training example must have a different subject/person performing the action. Fig. 8 (a) illustrates the Top-1, Top-3, Top-5 accuracy observed for each topology. We also report the average number of tunable parameters in each case. The reservoir has fixed connectivity from the input to reservoir as well as within the reservoir that are not affected during training. Hence, we only compare the total number of trainable parameters (comprising the weights from reservoir to output layer) as the reservoir size varies in Fig. 8 (a). We observe that the reservoir with 8K neurons yields maximum accuracy ($Top-1: 80.2\%, Top-5: 86.1\%$). Also, the classification accuracy improves as the reservoir size increases from 4K $\rightarrow$ 8K neurons. This is apparent since larger reservoirs with sparse random connectivity generate richer and heterogeneous dynamics that overall boosts the decoding capability of the network. Thus, for the same task difficulty (in this case 101-class problem), increasing the reservoir size improves the performance. However, beyond a certain point, the accuracy saturates. Here, the reservoir with 9K neurons achieves similar accuracy as 8K. In fact, there is a slight drop in accuracy from 8K $\rightarrow$ 9K. This might be a result of some sort of overfitting phenomenon as the reservoir (with more parameters than necessary for a given task) might begin to assign, instead of generalizing, its dynamic activity toward a certain input. Note, all reservoir models discussed so far have 10\% connectivity. Hence, the reservoirs in Fig. 8 (a) have increasing number of connections per neuron ($1:400 \rightarrow 1:900$) as we increase the number of neurons in the reservoir from 4000 to 9000. Please refer to the supplementary information (see Supplementary Fig. S1, S2) that details out the confusion matrix for the 101-class problem with 8K neurons and 10\% connectivity.

\begin{figure}[h]
\centering
\includegraphics[width=\linewidth]{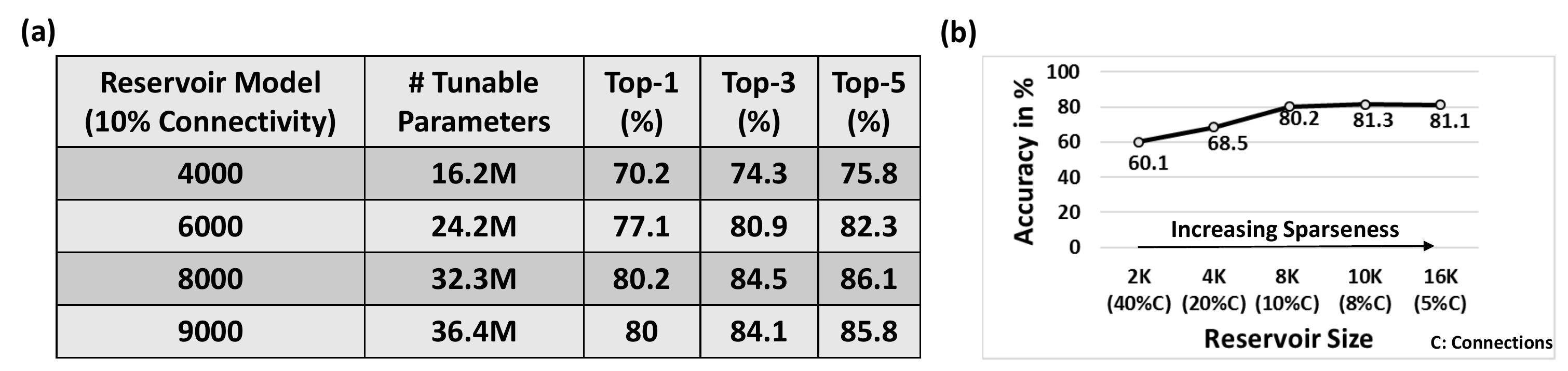}
\caption{(a) Top-1/3/5 accuracy on 101-classes obtained with 8-training example (per class) learning for D/A models of different topologies (b) Effect of variation in sparsity (for fixed $1:800$ number of connections per neuron across different reservoir topologies) on the Top-1 accuracy achieved by the reservoir. Note, all results are performed on the D/A reservoir scheme with 8 training videos per class processed with the input spike transformation technique mentioned earlier.}
\label{fig8}
\end{figure}

\subsection*{Impact of Variation of Reservoir Sparsity on Accuracy}
Since sparsity plays an important role in determining the reservoir dynamics, we simulated several topologies in an iso-connectivity scenario to see the effect of increasing sparseness on the reservoir performance. Fig. 8 (b) shows the variation in Top-1 accuracy for 101-classes as the reservoir sparseness changes. From Fig. 8 (a), we observe that the 8000N-10\% connectivity reservoir (with $1:800$ number of connections per neuron) gave us the best results. Hence, in the sparsity analysis, we fixed the number of connections per neuron to $1:800$ across different reservoir topologies. Initially, for a 2000N-40\% connectivity reservoir, the accuracy observed is pretty low ($60.1\%$). This is evident as lesser sparsity limits the richness in the complex dynamics that can be produced by the reservoir that further reduces its’ learning capability. Increasing the sparseness improves the accuracy as expected. We would like to note that the 4000N-20\% connectivity reservoir yields lesser accuracy than that of the sparser 4000N-10\% connectivity reservoir of Fig. 8 (a). Another interesting observation here is that the 10,000N-8\% reservoir yields higher accuracy ($\sim 1\%$ more) than that of the 8000N-10\%  reservoir. This result supports the fact that in an iso-connectivity scenario, more sparseness yields better results. However, the accuracy again saturates with a risk of overfitting or losing generalization capability beyond a certain level of sparseness as observed in Fig. 8 (b). Hence, there is no fixed rule to determine the exact connectivity/sparsity ratio for a given task. We empirically arrived at the results by simulating a varying set of reservoir topologies. However, in most cases, we observed that for an N-reservoir model 10\% connectivity generally yields reasonable results as seen from Fig. 8 (a). 
It is worth mentioning that while the 10K neuron reservoir yields the highest accuracy of $81.3\%$, it consists of larger number of tunable parameters (total connections/weights from reservoir to output) $40.4M$ as compared to $32.3M$ observed with the 8K-10\% connectivity reservoir (Fig. 8 (a)). In fact, the best accuracy achieved with our approach corresponds to the 10K-8\% connectivity reservoir with Top-1/Top-3/Top-5 accuracy as 81.3\%/85.2\%/87\%, respectively. Hence, depending upon the requirements, the connectivity/size of the reservoir can be varied to achieve a favorable efficiency-accuracy tradeoff. Please note that the total number of tunable parameters in the reservoir model include all 5-autonomous models.

\subsection*{EigenValue Spectra Analysis of Reservoir Connections}
We analysed the EigenValue (EV) spectra of the synaptic connections (both $J_{slow}$ and $J_{fast}$ of the autonomous models) to study the complex dynamics of the reservoir. EV spectra has been shown to characterize the memory/learning ability of a dynamical system \cite{rajan2009spontaneous}. Rajan et. al \cite{rajan2009spontaneous, rajan2006eigenvalue} have shown that stimulating a reservoir with random recurrent connections activates multiple modes where neurons start oscillating with individual frequencies. These modes then superpose in a highly non-linear manner resulting in the complex chaotic/ persistent activity observed from a reservoir. The EVs of the synaptic matrix typically encode the oscillatory behavior of each neuron in the reservoir. The real part of an EV corresponds to the decay rate of the associated neuron, while frequency is given by the imaginary part. If the real part of a complex EV exceeds 1 (or Re(EV)) $>$ 1), the activated mode leads to long lasting oscillatory behavior. If there are a few modes with Re(EV)) $>$ 1 , the network is at a fixed point, meaning the reservoir exhibits the same pattern of activity for a given input. It is desirable to construct a reservoir with multiple fixed points so that each one can be used to retain a different memory. In a recognition scenario, each of the fixed points latch onto a particular input pattern or video in our case, thereby yielding a good memory model. As the number of modes with Re(EV) $>$ 1 increases, the reservoir starts generating chaotic patterns, that are complex and non-repeating. In order to construct a reservoir model with good memory, we must operate in a region between singular fixed point activity (Re(EV) $<<$ 1) and complete chaos (Re(EV) $>>$ 1). 

\begin{figure}[h]
\centering
\includegraphics[width=\linewidth]{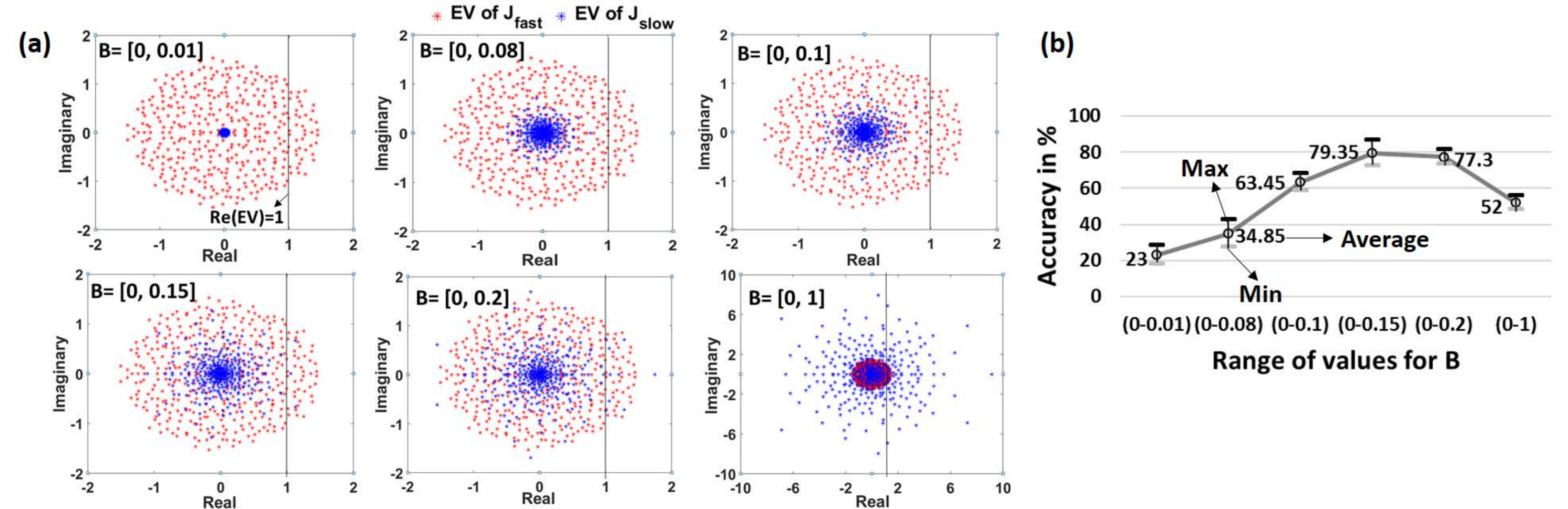}
\caption{(a) Change in EigenValue spectra of the $J_{fast}$ and $J_{slow}$ connections of an autonomous network . Note, the $J_{fast}$ spectra (corresponding to the connections imported from driven model) remains constant, while the $J_{slow}$ spectra changes with varying $B$ values. (b) Impact of variation in $B$, representative of stable/chaotic reservoir dynamics, on the accuracy for 2-class (BenchPress/GolfSwing) 1-training example problem (Fig. 6). The minimum and maximum accuracy observed across 5 different trials, where each trial uses a different video for training, is also shown.}
\label{fig9}
\end{figure}

Since our autonomous models have two kinds of synapses with different time constants, we observe two different kinds of EV spectra that contribute concurrently to the dynamical state of the auto network. While the overall reservoir dynamics are dictated by the effective contribution of both connections, $J_{slow}$ (due to larger time constant of synaptic decay) will have a slightly more dominating effect. Fig. 9 (a) illustrates the EV spectra corresponding to the bottom scan autonomous 800N-reservoir model simulated earlier for the 2-class problem (BenchPress/GolfSwing) with just 1 training video (refer Fig. 6). Now, the EV of $J_{fast}$ have considerable number of modes with Re(EV) $>$ 1 that is characteristic of chaotic dynamics. Hence, as discussed above, we need to compensate for the overwhelming chaotic activity with modes that exhibit fixed points to obtain a reliable memory model. This can be attained by adjusting the $J_{slow}$ connections. From Fig. 9 (a), we see that as the range of $B$ (random constant used to derive the $J_{slow}$ connections from the driven model) increases, the EV spectral radius of the $J_{slow}$ connections grows. For $B = [0-0.01]$, the EVs are concentrated at the center. Such networks have almost no learning capability as the dynamics converge to a single point that produces the same time-independent pattern of activity for any input stimulus. As the spectral circle enlarges, the number of modes with Re(EV) $>$ 1 also increases. For  $B = [0-0.15]$, the number of fixed point modes reaches a reasonable number that in coherence with the chaotic activity due to $J_{fast}$ results in generalized reservoir learning for variable inputs. Increasing B further expands the spectral radius for $J_{slow}$ (with more Re(EV) $> $1) that adds onto the chaotic activity generated due to $J_{fast}$. In such scenarios, the reservoir will not be able to discern between inputs that look similar due to persistent chaos. This is akin to the observations  about the need for the reservoir to strike a balance between pattern approximation and separability \cite{rajan2009spontaneous}.

Fig. 9 (b) demonstrates the accuracy variation for the same 2-class 1-training example learning model with different values of B. We repeated the experiment 5 times wherein we chose a different training video per class in each case. As expected, the accuracy for a fixed state reservoir with $B = [0-0.01]$ is very low. As the number of modes increase, the accuracy improves justifying the relevance of stable and chaotic activity for better learning. As the chaotic activity begins to dominate with increasing B, the accuracy starts declining. We also observe that the overall accuracy trend (for the 2-class 1-training example model) represented by the min-max accuracy bar is consistent even when the training videos are varied. This indicates that the reservoir accuracy is sensistive to the B-initialization and is not biased towards any particular training video in our limited training example learning scheme.  While the empirical result corroborates our theoretical analysis and justifies the effectiveness of using the D/A approach in reservoir model construction, further investigation needs to be done to gauge the other advantages/disadvantages of the proposed approach. Please refer to \cite{rajan2006eigenvalue, rajan2009spontaneous} for more clarification on random matrices and EV spectra implication on dynamical reservoir models.

\section*{Comparison with State-of-the-Art Recognition Models}
Deep Learning Networks (DLNs) are the current state-of-the-art learning models that have made major advances in several recognition tasks \cite{cox2014neural, mnih2013playing}. While these large-scale networks are very powerful, they inevitably require large training data and enormous computational resources. Spiking networks offer an alternative solution by exploiting event-based data-driven computations that makes them attractive for deployment on real-time neuromorphic hardware where power consumption and speed become vital constraints \cite{han2015learning, merolla2014million}. While research efforts in spiking models have soared in the recent past \cite{maass2016energy, masquelier2007unsupervised, srinivasa2012self, panda2016unsupervised}, the performance of such models are not as accurate as compared to their artificial counterparts (or DLNs). From our perspective, this work provides a new standard establishing the effectiveness of spiking models and their inherent timing-based processing for action recognition in videos. 

Fig. 10 compares our reservoir model to state-of-the-art recognition results \cite{simonyan2014two, srivastava2015unsupervised, feichtenhofer2016convolutional}. We simulated a VGG-16 \cite{simonyan2014very} DLN architecture that only processes the spatial aspects of the video input. That is, we ignore the temporal features of the video data and attempt to classify each clip by looking at a single frame. We observe that the spatial VGG-16 model yields reduced accuracy of $66.2\%$ ($42.9\%$) when presented with the full training dataset (8-training examples per class), implying the detrimental effect of disregarding temporal statistics in the data. Please refer to the supplementary information (see Supplementary Fig. S3) that details out the efficiency vs. accuracy tradeoff between our proposed reservoir and the spatial VGG-16 model. The static hierarchical/feedforward nature of DLNs serves as a major drawback for sequential (such as video) data processing. While deep networks extract good flexible representations from static images, addition of temporal dimension in the input necessitates the model to incorporate recurrent connections that can capture the underlying temporal correlation from the data. 

Recent efforts have integrated Recurrent Neural Networks (RNNs) and Long Short-Term Memory (LSTMs) with DLNs to process temporal information \cite{srivastava2015unsupervised, donahue2015long}. In fact, the state-of-the-art accuracy on UCF-101 is reported as $92.5\%$ \cite{feichtenhofer2016convolutional} as illustrated in Fig. 10, where a two-stream convolutional architecture with two VGG-16 models, one for spatial and another for temporal information processing is used. While the spatial VGG model is trained on the real-valued RGB data, the temporal VGG model uses optical flow stacking to capture the correlated motion. Here, both the models are trained on the entire UCF-101 dataset that significantly increases the overall training complexity. All models noted in Fig. 10 (including the LSTM model) use the temporal information in the input data by computing the flow features and subsequently training the deep learning/ recurrent models on such inputs. The overall accuracy shown in Fig. 10  is the combined prediction obtained from the spatial RGB and temporal flow models. The incorporation of temporal information drastically improves a DLN's accuracy. 

Fig. 10 also quantifies the benefits with our reservoir approach as compared to the state-of-the-art models from a computational efficiency perspective. We quantify efficiency in terms of total number of resources utilized during training a particular model. The overall resource utilization is defined as the product of the Number of (tunable) Parameters $\times$ Number of training Data Examples. The number of data examples are calculated as Number of Frames per video $\times$ Number of Training videos ($\equiv 180 \times 9537$ for full training data). Referring to the results for spatial VGG-16 model, we observe that the reservoir (10K-10\% connectivity) is $2.4\times$ more efficient, while yielding significantly higher accuracy, in the iso-data scenario where both the models are shown 8 video samples per class. In contrast, the efficiency improves extensively to $28.5\times$ in the iso-accuracy scenario wherein the reservoir still uses 8 video samples per class (808 total videos), while the VGG-16 model requires the entire training dataset (9537 total videos) during training. Comparing with the models that include temporal information, our reservoir (10K-10\% connectivity), although with lower accuracy, continues to yield significantly higher benefits of approximately $54 \times$ and $7\times$ as compared to Two-stream architecture and Temporal Convolutional network, respectively, on account of being trained on limited data. However, in limited training example scenario, the reservoir (81.3\% accuracy) continues to outperform an LSTM based machine learning model (67\% accuracy) while requiring lesser training data. It is worth mentioning that our reservoir due to the inherent temporal processing capability requires only 30 epochs for training. On the other hand, the spatial VGG-16 model needs 250 epochs. As the number of epochs required for training increase, the overall training cost (can be defined as, Number of Epochs $\times$ Resource Utilization) would consequently increase. This further ascertains the potential of reservoir computing for temporal data processing.

\begin{figure}[h]
\centering
\includegraphics[width=0.65\linewidth]{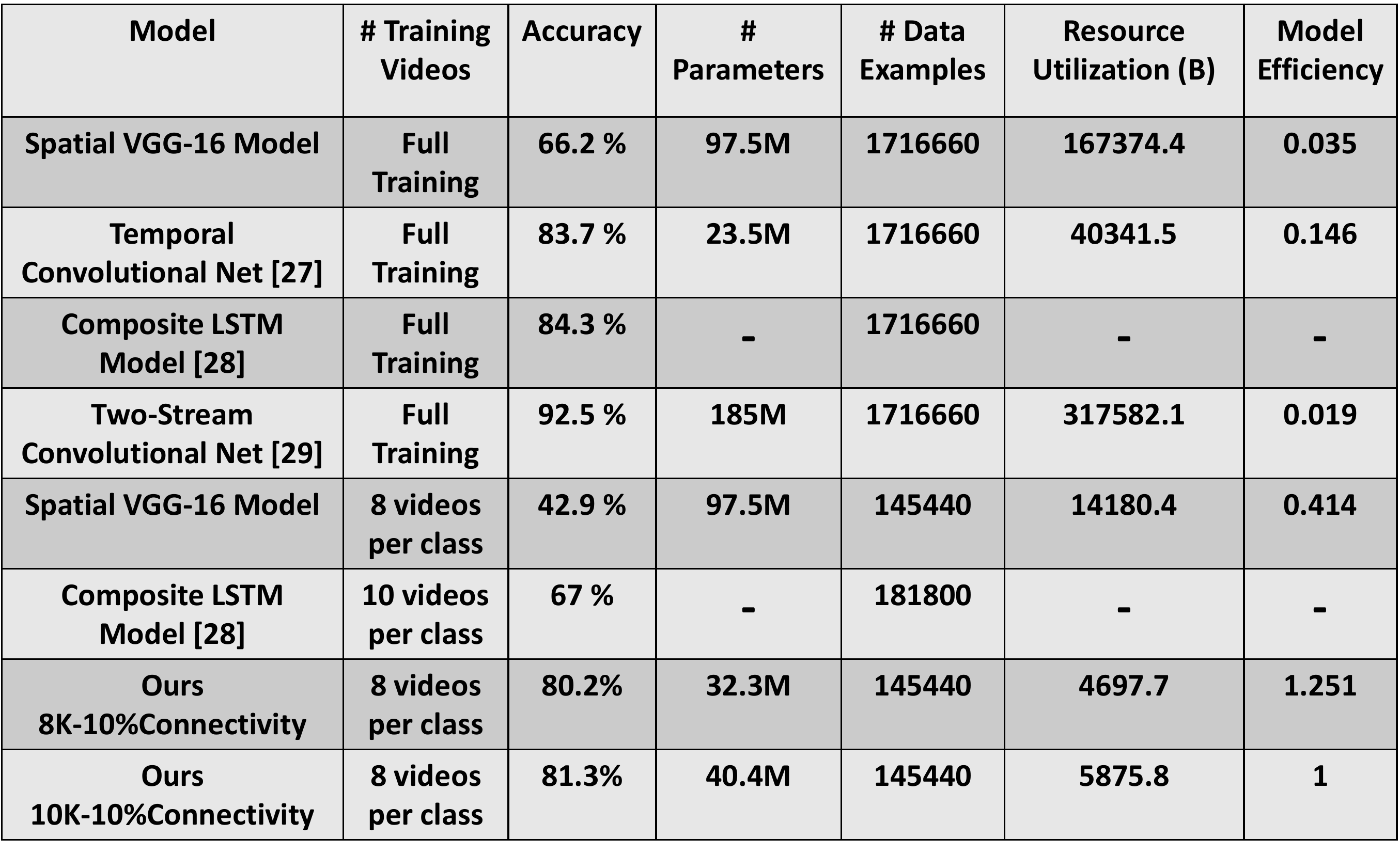}
\caption{Accuracy and Efficiency comparison with state-of-the-art recognition models. Resource Utilization (B) is the product of Number of Parameters $\times$ Number of Data Examples, where $B$ denotes billion. The model efficiency is calculated by normalizing the net resource utilization of each model against our reservoir model (10K-10\% connectivity). Full Training (in column corresponding to  \# Training Videos) denotes the entire training dataset that comprises of 9537 total videos across 101 classes.}
\label{fig10}
\end{figure}


\section*{Discussion}
In this paper, we demonstrated the effectiveness of a spiking-based reservoir computing architecture for learning from limited video examples on UCF101 action recognition task. We adopted the Driven/Autonomous reservoir construction approach originally proposed for neuroscientific use and gave it a new dimension to perform multi input-output transformation for practical recognition tasks. In fact, the D/A approach opens up novel possibilities for multi-modal recognition. For instance, consider a 5-class face recognition scenario. Here, the driven network is trained to produce 5 different desired activity. The auto models derived from the driven network will process the facial data to perform classification. Now, the same driven model can also be used to derive autonomous models that will work on, say voice data. Combining the output results across an ensemble of autonomous models will boost the overall confidence of classification. Apart from performance, the D/A model also has several hardware oriented advantages. Besides sparse power-efficient spike communication, the reservoir internal topology across all autonomous models for a particular problem remains equivalent. This provides a huge advantage of reusability of network structures while scaling from one task to another. 

Our analysis using eigenvalue spectra results on the reservoir provide a key insight about D/A based reservoir construction such that the complex dynamics can converge to fixed memory states. We believe that this internal stabilization enables it to generalize its memory from a few action signatures. We also compared our reservoir framework with state-of-the-art models and reported the cost/accuracy benefits gained from our model. We would like to note that our work delivers a new benchmark for action recognition from limited training videos in spiking scenario. We believe that we can further improve the accuracy with our proposed approach by employing better input encoding techniques. The scan-based filtering approach discussed in this paper does not account for depth variations and is susceptible to extreme spatial/rotational variation. In fact, we observe that our reservoir model yields higher accuracy for classes with more consistent action signatures and finds it difficult to recognize classes with more variations in movement (refer to Supplementary Fig. S1 for action specific accuracies). Using an adequate spike processing technique (encompassing depth based filtering, motion based tracking or a fusion of these) will be key to achieving improved accuracy. Another intriguing possibility is to combine our spike based model that can rapidly recognize simpler and repeatable actions (e.g., GolfSwing) with minimal samples for training with more other DLN models for those actions that have several variations (e.g., Knitting) or the actions that appear different dependent on the view point or both. We are also studying the extension of our SNN model to a hierarchical model that could possibly capture more complex and subtle aspects of the spatio-temporal signatures in these actions. 

Finally, our reservoir model, although biologically inspired, only has excitatory components. The brain’s recurrent circuitry consists of both excitatory and inhibitory neurons that operate in a balanced regime giving it the staggering ability to make sense of a complex and ever-changing world in the most energy-efficient manner \cite{herculano2012remarkable}. In the future, we will concentrate on incorporating the inhibitory scheme that will further augment the sparseness in reservoir activity thereby improving the overall performance. An interesting line of research in this regard can be found here \cite{srinivasa2014unsupervised, brendel2017learning}. Although there have been multiple efforts in the spiking domain exploring hierarchical feedforward and reservoir architectures with biologically plausible notions, most of them have been focused on static images (and some voice recognition) that use the entire training dataset to yield reasonable results \cite{kheradpisheh2016bio, diehl2015fast}. To that effect, this work provides a new perspective on the power of temporal processing for rapid recognition in cognitive applications using spiking networks.  In fact, we are currently exploring the use of this model for learning at the edge (such as cell phones and sensors where energy is of paramount importance) using a recently developed neural chip called the Loihi \cite{johnintel} that enables energy efficient on-chip spike timing based learning. We believe that results from this work could be leveraged to enable Loihi based applications such as video analytics, video annotation and rapid searches for actions in video databases.

\section*{Acknowledgements}
We would like to acknowledge the support of Priyadarshini Panda through a summer internship in the Microarchitecture Research Labs at Intel Corporation. We would also like to thank Michael Mayberry and Daniel Hammerstrom for their helpful comments and suggestions.


\section*{\fontsize{16}{16}\selectfont Supplementary Information}

Fig. S1 shows the class-wise accuracy obtained with our reservoir approach. We can deduce that certain classes are learnt more accurately than the others. This is generally seen for those classes that have more consistent action signatures and lesser variation such as \textit{GolfSwing, PommelHorse} among others. On the other hand, classes that have more variations in either action or views or both, such as \textit{TennisSwing, CliffDiving} etc yield less accuracy. In fact, classes that consist of many moving subjects such as \textit{MilitaryParade, HorseRacing} exhibit lower accuracy. In such cases, the input processing technique may not be able to extract motions from all relevant moving subjects. Here, irrelevant clutter or jitter due to camera motion gets captured in the spiking information as a result of which the reservoir is unable to recognize a particular signature. Such limitations can be eliminated with an adequate processing technique that incorporates depth and (or) motion based tracking. Also, data augmentation with simple transformations can further improve the accuracy. For instance, for a given training video, incorporating both left/right handed views by applying a 180 $\deg$ rotational transformation to all the frames of the original video will yield a different view/action for the same video. Such data augmentation will account for more variations in views/angles enabling the reservoir to learn better. 

Fig. S2 shows the confusion matrix of the 101-class problem for the same reservoir topology as Fig. S1.

\begin{figure}[h!]
\centering
\includegraphics[width = \textwidth]{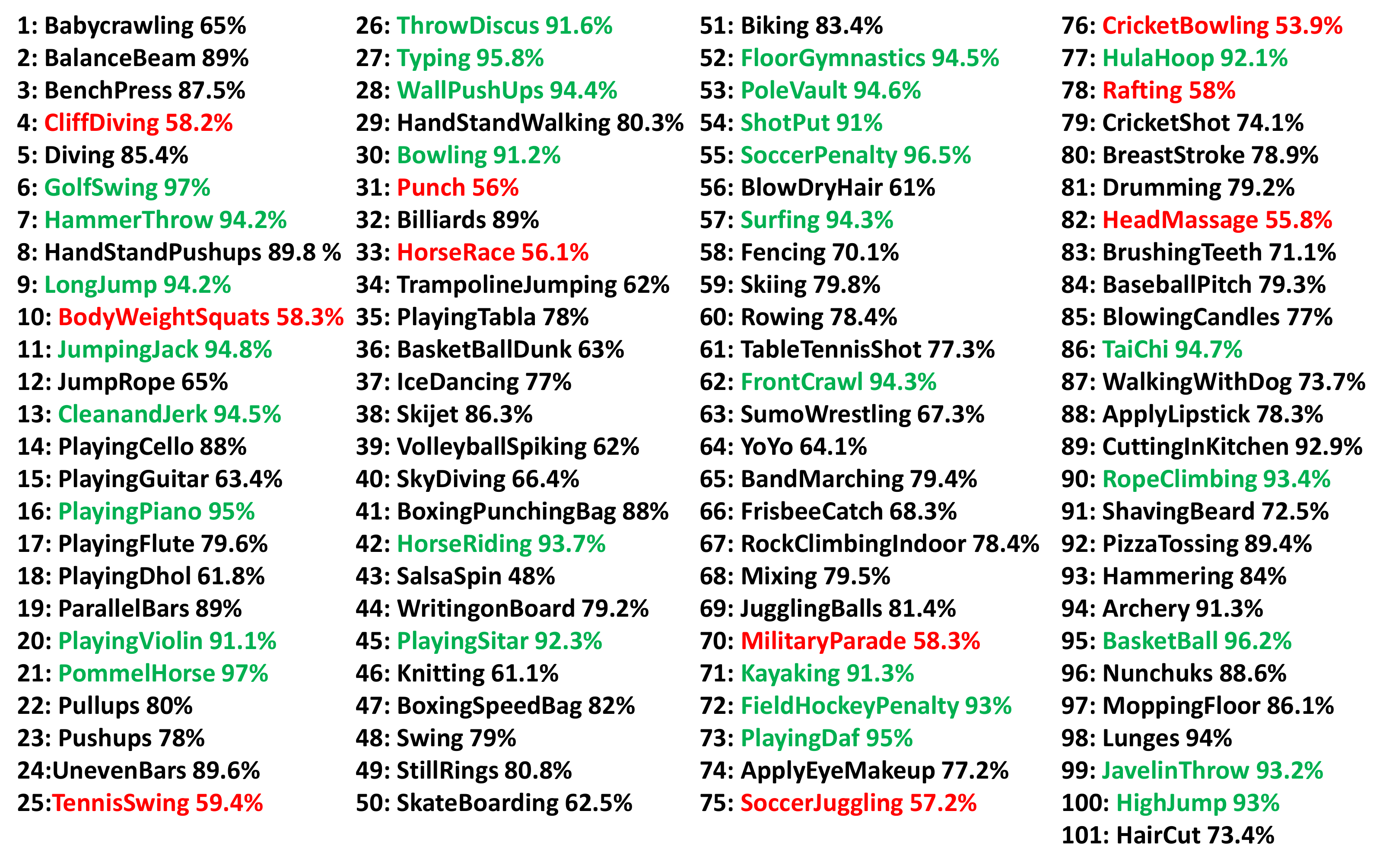}
\captionsetup{labelformat=empty}
\caption{\textbf{Figure S1.} Class-wise accuracy obtained with 8000 neuron (10\% connectivity) model on UCF-101 dataset when trained with 8 training videos per class. Few select classes with high/low accuracy are highlighted with green/red font to quantify the classes with more/less consistent videos.}
\label{fig1}
\end{figure}

\begin{figure}[h!]
\centering
\includegraphics[width=\textwidth]{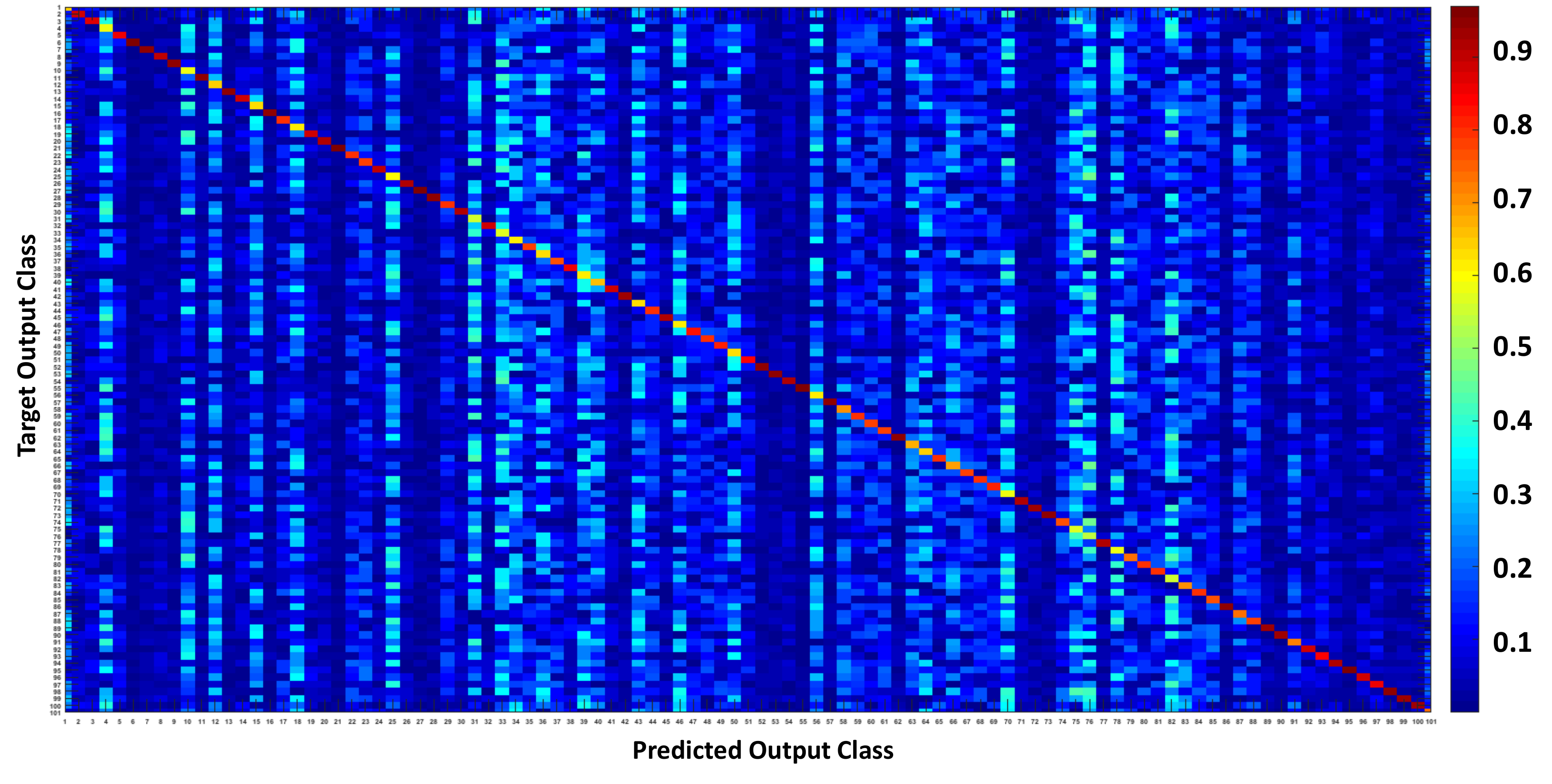}
\captionsetup{labelformat=empty}
\caption{\textbf{Figure S2.} Confusion Matrix for 101-class problem corresponding to Fig. S1 above. The X/Y axis represent the class numbers with equivalent notations as that of Fig. S1. The colormap indicates the accuracy for a given class label with highest intensity (red) corresponding to 100\% match between target \& output class and lowest intensity (blue) corresponding to 0\% match between target \& output class.}
\label{fig2}
\end{figure}

Here, we further quantify the advantages with our reservoir approach against the spatial VGG-16 model described in the main manuscript by gauging the efficiency vs. accuracy tradeoff obtained from both the models in equivalent training scenarios. Fig. S3 (a) illustrates the accuracy obtained for different training scenarios. For scenario 1, wherein the entire training dataset is used, the Top-1 accuracy obtained is $66.2\%$, thus implying the detrimental effect of disregarding temporal statistics in the data. Then, we tested the VGG model for two different limited example scenarios, one equivalent to our reservoir learning with 8 training examples per class and other with 40 training examples. In both cases, we see that the accuracy drops drastically with the 8-training example scenario yielding merely $42.9\%$ accuracy. Increasing the number of training examples to 40 does improve the accuracy to $50.02\%$, but the performance is much lower than that of the reservoir model. 

A noteworthy observation in Fig. S3 (a) is that the Top-5 accuracy for the VGG-16 model (89.8\%) trained with the entire dataset and the 8000N-10\% connectivity reservoir (86.1\%) trained on 8 videos per class (refer to Fig. 8 (a) in main manuscript) is comparable. In contrast, for limited training scenarios, the Top-5 accuracy with VGG is drastically reduced as compared to the reservoir models shown in Fig. 8 (a) of the main manuscript. This is indicative of the DLN's inferior generalization capability on temporal data. 

Fig. S3 (b) shows the normalized benefits observed with the different topologies of reservoir (Fig. 8 (a) in main manuscript) as compared to spatial VGG-16. Here, we quantify efficiency in terms of total number of trainable parameters (or weights) in a given model. The VGG-16 model for 101-class has 97.5M tunable parameters, against which all other models in Fig. S3 (b) are normalized. We observe $\sim6\times$ improvement in efficiency with the reservoir model where both VGG and the reservoir (\textit{4000-N reservoir with 20\% connectivity}) have almost similar accuracy $\sim67\%$. A noteworthy observation here is that VGG-16 in this case was trained with the entire UCF101 training data, while our reservoir had only 8 samples per class for training. This ascertains the potential of reservoir computing for temporal data processing. In an iso-data scenario (i.e. 8 videos per class for both VGG and reservoir models), we observe that the reservoirs yielding maximal accuracy of $81.3\% (80.2\%)$ continue to be more efficient by $2.2\times (3\times)$ than the VGG model. 

\begin{figure}[h!]
\centering
\includegraphics[width=\textwidth]{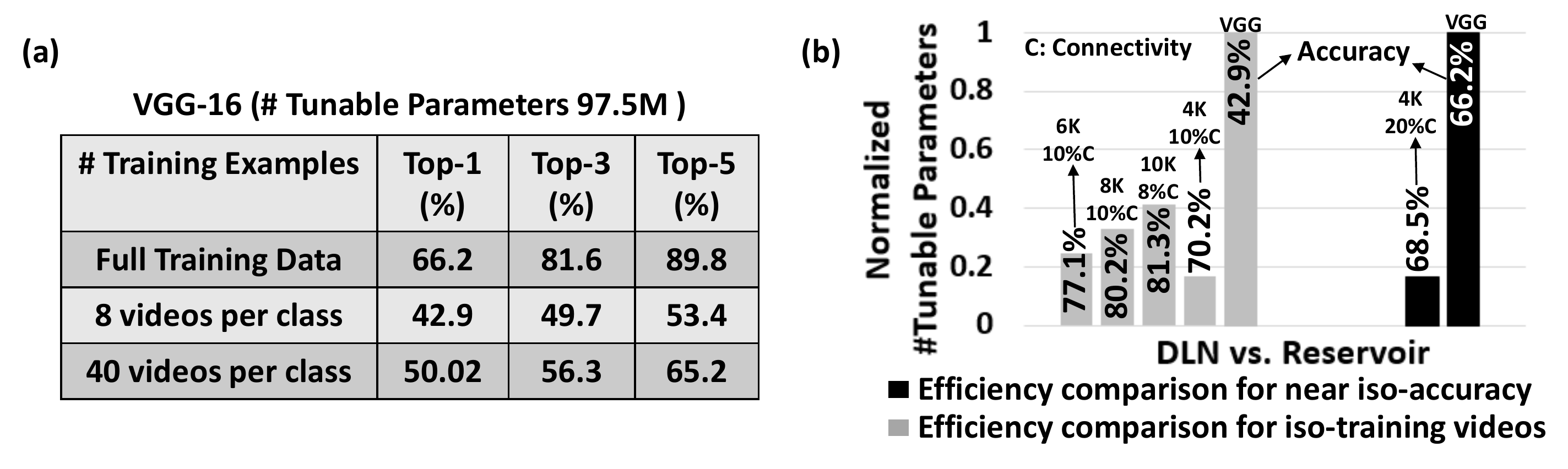}
\captionsetup[labelformat=empty]
\caption{\textbf{Figure S3.} (a) Top-1/3/5 accuracy obtained with the VGG-16 deep learning model on 101-classes for different training scenarios (b) Efficiency (quantified as total \# of tunable parameters) comparison between VGG-16 model and D/A reservoir model of different topologies (from Fig. 8 (a) in main manuscript) for iso-accuracy and iso-data (same number of training examples) scenario. The accuracy/topology for each model is noted on the graph.}
\label{fig3}
\end{figure}

For implementing the VGG-16 convolutional network model, we used the machine learning tool, Torch \cite{collobert2011torch7}. Also, we used standard regularization Dropout \cite{srivastava2014dropout} technique to optimize the training process for yielding maximum performance. In case of the DLN model, we feed the raw pixel valued RGB data as inputs to the DLN for frame-by-frame training thereby providing the model with all information about the data. The input video frames originally of ${200\times300}$ dimensions are resized to ${224\times224}$. The macro-architecture of the VGG-16 model was imported from \cite{sergey} and then altered for 101-class classification. To ensure sufficient number of training examples for the deep VGG-16 model in the limited example learning cases, we augmented the data via two random transformation schemes: rotation in the range of $(0-40 \deg )$, width/ height shifts within $20 \%$ of total width/height.  For the 8/40-training example scenario, each frame of a training video was augmented 10/2 times, respectively using any of the above transformations. The VGG-16 model in each of the training example scenarios described above is trained for 250 epochs. Here, in each epoch, the entire limited/full training dataset (that is the individual frames corresponding to all classes) are presented to the VGG model. Note, UCF101 provides three train/test split recommendations. We use just split \#1 for all of our experiments.

\end{document}